\documentclass{article}

\usepackage[preprint]{neurips_2025}

\usepackage[utf8]{inputenc} % allow utf-8 input
\usepackage[T1]{fontenc}    % use 8-bit T1 fonts
\usepackage{hyperref}       % hyperlinks
\usepackage{url}            % simple URL typesetting
\usepackage{booktabs}       % professional-quality tables
\usepackage{amsfonts}       % blackboard math symbols
\usepackage{nicefrac}       % compact symbols for 1/2, etc.
\usepackage{microtype}      % microtypography

\usepackage{wrapfig}
\usepackage{amsmath}
\usepackage{graphicx}

\usepackage[frozencache,cachedir=minted-cache]{minted}
\usepackage{times}
\usepackage{soul}
\usepackage{url}
\usepackage{hyperref}
\usepackage[utf8]{inputenc}
\usepackage{caption}
\usepackage{graphicx}
\usepackage{amsmath}
\usepackage{amsthm}
\usepackage{booktabs}
\usepackage{algorithm}
\usepackage{algorithmic}
\usepackage{tcolorbox}
\usepackage{hyperref}
\hypersetup{hidelinks=true}
\usepackage{textcomp}
\usepackage{multirow} 
\usepackage{multicol} 
\usepackage{booktabs}
\usepackage{makecell}   % 引入 makecell 宏包，用于单元格内换行和表头格式

\usepackage{array}
\usepackage{booktabs} % 如果需要高级表格功能
\usepackage{subcaption} % 用于插入子图
\usepackage{caption}
\usepackage{enumitem}

\newcommand{\emb}[1]{\texttt{{\color{myblue}{#1}}}}
\newcommand{\newemb}[1]{\texttt{{\color{myblue2}{#1}}}}
\definecolor{myblue}{HTML}{CF553D}
\definecolor{myblue2}{HTML}{4472C4}
\definecolor{myblue3}{HTML}{ffff99}

\usepackage[utf8]{inputenc} % allow utf-8 input
\usepackage[T1]{fontenc}    % use 8-bit T1 fonts

\usepackage{amsfonts}       % blackboard math symbols
\usepackage{nicefrac}       % compact symbols for 1/2, etc.
\usepackage{microtype}      % microtypography

\title{ModuLM: Enabling Modular and Multimodal Molecular Relational Learning with Large Language Models}

\author{%
Zhuo Chen$^{1,2}$, \textbf{Yizhen Zheng}$^{4}$, \textbf{Huan Yee Koh}$^{4}$, \textbf{Hongxin Xiang}$^3$, \textbf{Linjiang Chen}$^5$, \\ 
\textbf{Wenjie Du}$^{1,2}$, \textbf{Yang Wang}$^{1,2}$\\  % Using * and † for corresponding author
$^1$University of Science and Technology of China, China\\
$^2$Suzhou Institute for Advanced Research, USTC, China\\
$^3$Hunan University, China \quad 
$^4$Monash University, Australia\\
$^5$State Key Laboratory of Precision and Intelligent Chemistry, USTC, China\\ 
\texttt{\{czchenzhuo, duwenjie\}@mail.ustc.edu.cn}\\
\texttt{\{yizhen.zheng1, huan.koh\}@monash.edu}\\
\texttt{xianghx@hnu.edu.cn} \quad
\texttt{\{linjiangchen, angyan\}@ustc.edu.cn}
}

\begin{document}

\maketitle

\begin{abstract}
Molecular Relational Learning (MRL) aims to understand interactions between molecular pairs, playing a critical role in advancing biochemical research. With the recent development of large language models (LLMs), a growing number of studies have explored the integration of MRL with LLMs and achieved promising results. However, the increasing availability of diverse LLMs and molecular structure encoders has significantly expanded the model space, presenting major challenges for benchmarking. Currently, there is no LLM framework that supports both flexible molecular input formats and dynamic architectural switching. To address these challenges, reduce redundant coding, and ensure fair model comparison, we propose ModuLM, a framework designed to support flexible LLM-based model construction and diverse molecular representations. ModuLM provides a rich suite of modular components, including 8 types of 2D molecular graph encoders, 11 types of 3D molecular conformation encoders, 7 types of interaction layers, and 7 mainstream LLM backbones. Owing to its highly flexible model assembly mechanism, ModuLM enables the dynamic construction of over 50,000 distinct model configurations. In addition, we provide comprehensive  results to demonstrate the effectiveness of ModuLM in supporting LLM-based MRL tasks. ModuLM is available at~\url{https://anonymous.4open.science/r/ModuLM}.
\end{abstract}
% https://github.com/ssjjjhw/ModuLM
% https://anonymous.4open.science/r/ModuLM

\section{Introduction}
Molecular Relational Learning (MRL) \citep{CGIB}, which aims to understand the interactions between molecular pairs, has garnered growing attention due to its wide-ranging applications across various scientific domains \cite{drug_drug_3}. For example, drug-drug interactions (DDIs) are vital for understanding the effects of concurrent drug use, which can inform strategies to prevent adverse drug reactions and ensure patient safety \cite{drug_drug_1}, while solute-solvent interactions (SSIs) are fundamental to solution chemistry and are pivotal in the design and optimization of chemical processes \cite{SSI_1,SSI_2}. However, the exhaustive experimental validation of these interactions is notoriously time-consuming and costly.

In recent years, large language models (LLMs) have emerged as a promising new paradigm in MRL research due to their powerful capabilities in knowledge integration and reasoning. Compared with traditional methods, LLMs can more efficiently process and understand complex interactions between molecules, significantly improving modeling performance and generalizability. A growing body of research has focused on LLM-based MRL frameworks \cite{LLM4MI1,LLM4MI2,fang2024moltc},  leveraging the strengths of LLMs to achieve strong results on MRL tasks. For instance, ReactionT5\cite{LLM4MI4} proposed a text-based pretrained LLM tailored for MRL tasks, while MolTC\cite{fang2024moltc} further advanced this line of research by integrating multimodal data and incorporating 2D molecular graphs for improved performance. These developments highlight the research value and application potential of LLMs in MRL. However, with the emergence of an increasing number of encoding methods and backbone models~\cite{Nguyen_Le_Quinn_Nguyen_Le_Venkatesh_2020, Huang_Xiao_Glass_Sun_2020, Voitsitskyi__2023, Zhu_Zhao_Wen_Wang_Wang_2023, Yazdani-Jahromi_2022b, taylor2022galactica, touvron2023llama, guo2025deepseek}, it is now possible to adopt more flexible strategies to recombine components and build more novel model architectures.  However, this flexibility introduces new challenges for the benchmarking and evaluation of LLM-based MRL models.

\textbf{Lack of diverse input support:} Molecular structures can typically be represented in various forms, such as 1D SMILES strings, 2D molecular graphs, and 3D molecular conformations; however, most existing models support only a single representation modality, commonly 1D SMILES\cite{kipf2016semi,gilmer2017neural,xu2018representation} or 2D graphs\cite{gasteiger2021gemnet,schutt2021equivariant,du2022se,du2023new}, which limits their ability to fully capture the complexity of molecular interactions and may result in the loss of critical structural information. In the field of LLMs, MRL models that accept 3D molecular structure inputs remain extremely rare, despite the fact that specific 3D conformations are often essential for accurately modeling chemical phenomena, for instance, in small-molecule binding to target proteins\cite{wu2022pre}. These issues highlight the importance of a unified framework that can accommodate 1D, 2D, and 3D molecular inputs to enable more comprehensive, flexible, and accurate molecular relational learning across a wide range of task scenarios.

% \textbf{Lack of Flexible Architectures:} Current LLM-based MRL models often adopt relatively rigid architectures. For example, ReactionT5 \cite{LLM4MI4} uses a unified model to encode both SMILES sequences and molecular property descriptions, while MolTC \cite{fang2024moltc} employs graph neural networks (GNNs) to encode molecular graphs. Although these approaches achieve certain performance gains, they still show limitations in handling multimodal molecular data. In contrast, non-LLM MRL methods often use more effective encoding strategies—for instance, MMGNN \cite{du2024mmgnn} employs interpretable GNN to extract key subgraphs for solvation free energy prediction, and Uni-Mol \cite{Uni-mol} introduces pre-trained SE(3) Transformer models specifically designed for molecular data. Moreover, many current LLM-based MRL models overlook molecular interaction modeling. These challenges underscore the importance of developing a model framework capable of seamlessly combining diverse encoding modules while effectively incorporating molecular interaction information.

\textbf{Lack of Flexible Architectures:} Current LLM-based MRL models often adopt relatively rigid architectures. For example, ReactionT5 \cite{LLM4MI4} uses a unified model to encode both SMILES sequences and molecular property descriptions, while MolTC \cite{fang2024moltc} employs graph neural networks (GNNs) to encode molecular graphs. Although these methods achieve performance improvements, they are still somewhat limited by their encoding strategies. In the non-LLM MRL domain, many more effective encoding strategies have been developed, such as MMGNN \cite{du2024mmgnn}, which employs interpretable GNNs to extract key subgraphs for solvation free energy prediction, and Uni-Mol \cite{Uni-mol}, which introduces pre-trained SE(3) Transformer models specifically designed for molecular data. However, integrating these encoding methods into existing LLM frameworks remains a challenge. Furthermore, most current LLM-based models overlook the modeling of molecular interaction features. These challenges highlight the importance of developing a model framework that can seamlessly combine diverse encoding modules while effectively incorporating molecular interaction information.

To this end, we propose \textbf{ModuLM}, a unified and extensible framework designed to overcome the limitations of existing LLM-based MRL approaches. ModuLM provides a highly flexible model construction mechanism that supports a wide range of molecular input formats, multimodal integration strategies, and diverse prompt designs. The framework accepts molecular representations in the form of 1D SMILES, 2D molecular graphs and 3D conformations. It 
 includes 8 types of 2D molecular graph encoders, 11 types of molecular conformation encoders, 7 types of interaction feature encoders, and 7 mainstream LLM backbones, along with specially designed prompt templates for integrating different types of molecular features. To enhance usability and extensibility, ModuLM adopts a modular interface design that allows users to flexibly assemble and extend models, supports incremental pretraining, and is capable of handling complex molecular interaction modeling tasks. ModuLM can generate over 50,000 distinct model configurations. We conduct comprehensive benchmark experiments on tasks such as  DDI, SSI and CSI. The results demonstrate that ModuLM performs remarkably well in constructing, evaluating, and comparing LLM-based MRL models. These findings highlight ModuLM’s strong potential to advance the development of MRL models and provide valuable insights into molecular interaction mechanisms.

\section{Related Work}
\textbf{Molecular Relational Learning:} MRL is critical for drug research, with machine learning offering a scalable alternative to costly experimental validation~\cite{lee2023conditional}. Early methods focused on GNNs~\cite{kipf2016semi,zhong2019graph,RN460,fu2020core}, such as Nyamabo et al.’s substructure-level interaction model using GAT and co-attention~\cite{Co-attention}, and Lee et al.’s CGIB~\cite{CGIB}, which applies the information bottleneck to extract key substructures. LLM-based methods have gained momentum. For instance, ReactionT5~\cite{LLM4MI7} enhances molecular understanding by integrating chemical structures with natural language. MolTC~\cite{fang2024moltc} further advances this direction by combining 2D molecular graph features with chain-of-thought reasoning to support complex molecular inference.

\textbf{LLMs in the Molecular Domain:} LLMs have been widely applied in 1D, 2D, and 3D molecular pattern learning. For 1D, methods like MolT5~\cite{MolT5} and KV-PLM~\cite{KV-PLM} tokenize SMILES strings for representation learning. In 2D, approaches such as Text2Mol~\cite{Text2Mol}, MolCA~\cite{MolCA}, and DrugChat~\cite{DrguChat} integrate molecular graphs with text encoders or LLMs. For 3D, MolLM~\cite{MolLM} and 3D-MoLM~\cite{3D-MoLM} incorporate spatial relationships via attention mechanisms and 3D encoders. In addition, LLMs have also found applications in MRL, such as ReactionT5~\cite{LLM4MI4} and MolTC~\cite{fang2024moltc}, which utilize multimodal data, including molecular graphs (2D), chemical properties, and SMILES (1D), for MRL. 

\textbf{Deep Learning Frameworks Specialized for MRL:} DeepPurpose is a user-friendly deep learning library for drug-target interaction prediction. It supports customized model training with 15 compound and protein encoders and over 50 neural architectures~\cite{Huang_Fu_Glass_Zitnik_Xiao_Sun_2020}. FlexMol is another toolkit designed for MRL, offering a variety of encoders and interaction layers that support sequence-based and graph-based representations of drugs and proteins~\cite{sizhe2024flexmol}.

\section{ModuLM}
In this section, we introduce ModuLM following the model training workflow for LLM-based MRL.
\subsection{Framework}
\begin{figure}[ht]  % 可以指定浮动的位置，例如[h]表示在当前位置
    \centering  % 图片居中显示
    \includegraphics[width=1\textwidth]{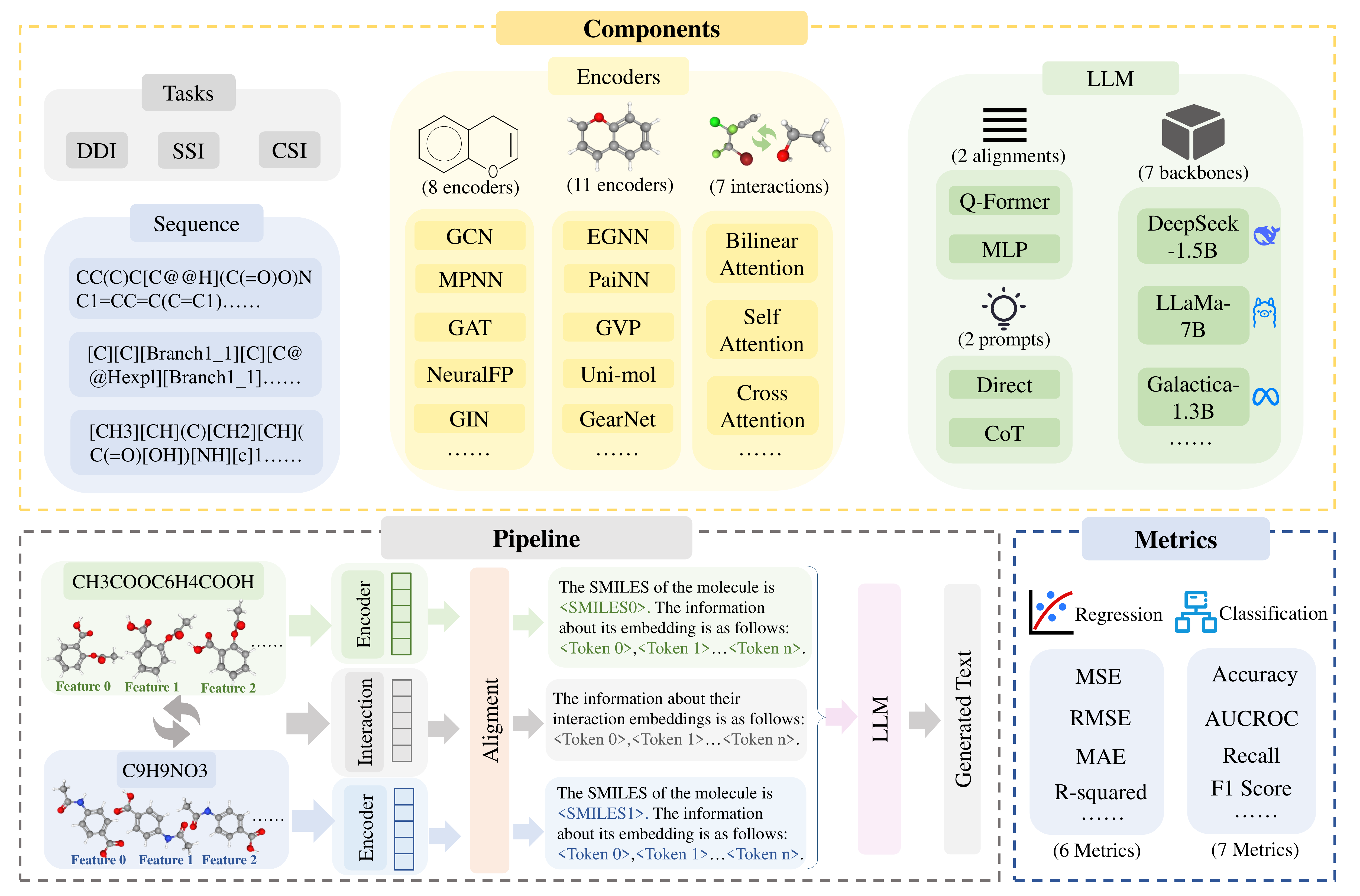}  % 这里设置图片的路径及尺寸
    \caption{Overview of the ModuLM framework.}  % 添加标题
    \label{fig:framework}  % 可选：为图片设置标签，方便在文中引用
\end{figure}
ModuLM is an LLM framework designed for MRL, with the overall architecture illustrated in Figure~\ref{fig:framework}. It supports three types of molecular text inputs and accommodates a variety of molecular inputs across 1D, 2D, and 3D modalities. The framework provides eight 2D molecular graph encoders, eleven 3D molecular conformation encoders, and incorporates seven feature interaction designs. It also supports evaluation across multiple task types.

\subsection{Post Pretraining}

\begin{table*}[ht]\centering
\begin{minipage}{\textwidth}\vspace{0mm}    \centering
\caption{Text settings for different pretraining methods.}
    \label{tab:prompt}
\begin{tcolorbox} 
    \centering
   
      \footnotesize
    \begin{tabular}{p{0.97\textwidth} c}
   \textbf{ {\bf Molecular Interaction-based Pretraining} } &\\
The first molecule has a SMILES representation of \emb{<SMILES0>}, which suggests certain structural characteristics and chemical functionalities. Based on its structure, it may exhibit the property \emb{[Property0]}, such as high solubility, bioactivity, or specific binding affinity. On the other hand, the second molecule is represented by the SMILES string \newemb{<SMILES1>}, and analysis of its structure indicates it may exhibit the property \newemb{[Property1]}, potentially contributing to its pharmacokinetic behavior or molecular interaction profile.

% \textbf{ {\bf Target Response for DDI Tasks (Fine-tuning)} } & \\
% Considering the molecular conformation, the first molecule may exhibit the property \emb{[Property0]}, while the second molecule may possess the property \texttt{[Property1]}. This interaction could potentially enhance the photoreactivity of the second molecule. Therefore, they are likely to interact with each other.

    \hrulefill & \\
    
     \textbf{ {\bf Substructure-based Pretraining} } &\\
The molecule has a SMILES representation of \emb{<SMILES0>}, which encodes its atomic connectivity and overall molecular structure. It contains notable substructures such as \emb{<substructe0>} and \emb{<substructe1>}, both of which are known to play significant roles in determining the molecule's physicochemical and biological properties. Based on the presence of these functional groups or structural motifs, the molecule is likely to exhibit the properties \emb{[Property0]} and \emb{[Property1]}, which may influence its reactivity, solubility, or interaction with biological targets.
 & \\

    \end{tabular}
\end{tcolorbox}
\vspace{-2mm}

\end{minipage}
\end{table*}

To strengthen the domain-specific capabilities of LLMs in chemistry, we begin with incremental pretraining to better adapt them for MRL tasks. In ModuLM, we conduct a survey of various authoritative biochemical databases, such as PubChem\footnote{\href{https://pubchem.ncbi.nlm.nih.gov}{https://pubchem.ncbi.nlm.nih.gov}} and DrugBank~\cite{PubChem}, collecting a large amount of molecular property description texts. We then provide three distinct pretraining strategies:  Molecular Interaction-based  
 pretraining, Substructure-based pretraining and Structure Similarity-guided 
 Grouping pretraining.

  \textbf{Molecular Interaction-based  Pretraining} is based on the pretraining method proposed by MolTC~\cite{fang2024moltc}. Its pretraining text setup is shown in the Table~\ref{tab:prompt}. Considering that molecular interaction tasks typically involve two different molecules, MolTC integrates the textual representations of both molecules and inputs them jointly during pretraining, enabling the LLM to develop prior knowledge in the form of molecular pairs.

   \textbf{Substructure-based  Pretraining} adopts a pretraining text setup as shown in the Table~\ref{tab:prompt}, enabling the LLM to learn more fine-grained information within molecules. This allows the LLM to make more accurate judgments based on the potential substructures of different molecules, thereby enhancing its generalization ability when encountering previously unseen molecules and improving performance on downstream tasks.

   \textbf{Structure Similarity-guided Grouping Pretraining} combines substructure-based pretraining with a grouping strategy based on structural similarity. Molecules with similar structures are grouped together and input as a group during pretraining, thereby enhancing the LLM’s understanding of this class of molecules. This method shares the same prompt configuration as the substructure-based pretraining approach, differing in the ordering of the input.

It is worth noting that if you use the Q-former approach for aligning textual and molecular data, additional pretraining will be required. The specific pretraining process can be selected based on your needs. Here, we provide MolTC's~\cite{fang2024moltc} methods for aligning molecules with text and molecular data.

\subsection{Fine-tuning}
\subsubsection{Input Data}
To transform raw molecules into meaningful representations, ModuLM first performs preprocessing followed by encoding. The preprocessing stage includes tasks such as tokenization, normalization, feature extraction, fingerprint generation, molecular graph construction, and molecular conformation generation. The encoding stage is responsible for dynamically constructing the input features for the LLM. In this stage, the preprocessed data is processed to generate embeddings that can be utilized by subsequent network layers. The following will present the specific procedures and methods used by ModuLM to encode different types of data.

\textbf{1D Representations of Molecules:}
The commonly used molecular representations for MRL  in current LLMs are typically SMILES and SelfIES~\cite{krenn2022selfies}. However, SMILES representations may have difficulties clearly expressing certain molecular structures. Therefore, for MRL tasks, we recommend using SelfIES. Nonetheless, to provide a more comprehensive benchmarking framework for previous LLM-based approaches, we include both molecular text encoding methods here, and additionally incorporate  SMARTS. For 1D representations of molecules, the encoding method depends on the specific LLM used.

\textbf{2D Molecular Graph Representation:} Since 1D molecular representations often fail to  capture the structural information of molecules, it is common in MRL tasks to incorporate multimodal molecular structure information. In non-LLM research areas, molecular graph representation is the most widely used approach. In most tasks, molecular graphs help models better understand molecular structures, thereby enhancing MRL performance. We provide a variety of methods as shown in the Table~\ref{tab:encode}.

\textbf{3D Molecular Conformation Representation:} In most MRL tasks, 2D molecular graphs are sufficient for the model to make effective judgments. However, in certain specialized tasks, it is necessary to incorporate the 3D spatial information of molecules to accurately determine intermolecular interactions. This is an aspect that current LLM-based MRL models have largely overlooked. In ModuLM, we address this limitation by providing various 3D structure encoding methods to support more advanced molecular relation learning. The encoding methods are listed in the Table~\ref{tab:encode}.

\textbf{Interaction:} 
In current LLM-based MRL model designs, the explicit modeling of molecular interaction relationships is often overlooked. In ModuLM, we address this issue by introducing specially designed prompts and interaction layers to incorporate interaction information into the LLM. The Interaction Layer is another key building block of the MRL model. These layers serve two main functions: capturing and modeling the relationships between different molecular entities, and integrating multiple embeddings of the same entity to form a more comprehensive feature representation. The Interaction Layer can accept encoded inputs from different molecules, enabling the construction of more complex model architectures. In ModuLM, the interaction designs we provide are shown in Table~\ref{tab:encode}.

\begin{table}[ht]
\centering
\caption{Encoding methods for different data formats}
\label{tab:encode}
\begin{tabular}{p{3cm}p{9cm}}
\toprule
\textbf{Encoder Type} & \textbf{Methods} \\
\midrule
\textbf{2D Graph} &  GCN \cite{kipf2017semi}, MPNN \cite{gilmer2017neural}, GAT \cite{velickovic2018graph}, NeuralFP \cite{duvenaud2015convolutional}, AttentiveFP \cite{Xiong_Wang_2019b}, GIN \cite{xu2019how}, GraphSAGE\cite{hamilton2017inductive}, CoATGIN\cite{zhang2022coatgin} \\
\midrule
\textbf{3D Conformation} & EGNN \cite{satorras2021n}, 3D-GeoFormer \cite{zhou2023self}, SE3Transformer \cite{fuchs2020se}, PaiNN \cite{schutt2021equivariant}, GVP \cite{jing2021learning}, GearNet \cite{zhang2022protein}, DimeNet++ \cite{gasteiger2020directional},  SchNet \cite{schutt2017schnet}, SphereNet \cite{liu2022spherical}, G-SphereNet\cite{luo2022autoregressive}, Uni-mol~\cite{zhou2023uni} \\
\midrule
\textbf{Interaction} & Bilinear Attention\cite{Bai_Miljković_John_Lu_2023}, Self Attention\cite{vaswani2017attention}, Cross Attention\cite{Qian_Li_Wu_Zhang_2023}, Highway\cite{Zhu_Zhao_Wen_Wang_Wang_2023}, Gated Fusion\cite{ren2018gated}, Bilinear Fusion\cite{lin2015bilinear}, Mean \\
\bottomrule
\end{tabular}
\end{table}

\subsubsection{Alignment}
Since data from different modalities typically exist in distinct semantic spaces, it is necessary to perform alignment before feeding them into the LLM. Currently, two common approaches are used: employing a lightweight MLP~\cite{liu2023visual} or using a Q-former~\cite{li2023blip}. In the ModuLM framework we provide, both alignment methods are supported. It is worth noting that when using a Q-former, it is typically involved during the pretraining stage to enable better alignment performance.

\subsubsection{Backbone}
Existing LLM-based MRL models often adopt different backbone architectures, and there has been no systematic investigation into the MRL performance across different LLM backbones. In ModuLM, we provide a streamlined method for switching backbones. We have surveyed and integrated a range of mainstream LLMs and offer simple interfaces for replacement, enabling easier comparison of performance differences across various types and scales of LLMs under a unified experimental setup. Here, we provide two types of prompts: direct inference and chain-of-thought-based reasoning. The specific prompt designs are detailed in the Appendix~\ref{app:prompt}.

\subsection{Evaluation Metrics}
ModuLM supports multiple default metrics, aligning with the TDC standard for molecular relational learning\cite{Huang2021TherapeuticsDC}. Users can specify the metrics in the Trainer for early stopping and testing. These metrics include various regression metrics (Mean Squared Error (MSE), Root-Mean Squared Error (RMSE), Mean Absolute Error (MAE), Coefficient of Determination (R\textsuperscript{2}), Pearson Correlation Coefficient (PCC), Spearman Correlation Coefficient), binary classification metrics (Area Under Receiver Operating Characteristic Curve (AUC-ROC), Area Under the Precision-Recall Curve (PR-AUC), Range LogAUC, Accuracy Metrics, Precision, Recall, F1 Score. 

\subsection{Supporting Datasets}
ModuLM is compatible with all MRL datasets that conform to our specified format. These datasets typically consist of three components: molecular entity one, molecular entity two, and a label. We provide utility functions to facilitate the loading of datasets in this format. In addition, ModuLM includes a wide range of built-in datasets from various domains, such as Drugbank (Version 5.0.3), ZhangDDI \cite{zhang2017predicting}, ChChMiner \cite{zitnik2018stanford}, DeepDDI \cite{ryu2018deep}, TWOSIDES \cite{tatonetti2012data}, Chromophore \cite{joung2020experimental}, MNSol \cite{marenich2020minnesota}, CompSol \cite{moine2017estimation}, Abraham \cite{grubbs2010mathematical}, CombiSolv \cite{vermeire2021transfer}, FreeSolv \cite{mobley2014freesolv}, and CombiSolv-QM \cite{vermeire2021transfer}. For more details, please refer to the Appendix~\ref{app:datasets}.

\section{Experiments}
We conduct validation experiments on MRL tasks using ModuLM to demonstrate the framework's capability in supporting a wide range of experiments, comparisons, and analyses. The following sections present the results for DDI  tasks, showcasing the impact of different inputs and encoders on the performance of LLMs in MRL. More experimental details are provided in the Appendix~\ref{app:exp_details}.
\subsection{Experimental Setup}
\label{wholeexp}
\begin{table*}[ht]
\caption{Experimental Settings on DDI Datasets}
\label{tab:experimental_settings}
\resizebox{\textwidth}{!}{
\begin{tabular}{c p{2.5cm} p{2.5cm} p{3cm} p{2.5cm}}
\toprule
\textbf{Experiment No.} & \textbf{Backbone} & \textbf{Encoder} & \textbf{Interaction} & \textbf{Input Feature} \\
\midrule
1.1 & Galactica-1.3B & - & - & \(m_s\) \\
1.2 & Galactica-1.3B & GIN & - & \(m_s + m_g\) \\
1.3 & Galactica-1.3B & GIN & Cross Attention & \(m_s + m_g\) \\
1.4 & Galactica-1.3B & Uni-mol & - & \(m_s + m_c\) \\
1.5 & Galactica-6.7B & MPNN & Gated Fusion & \(m_s + m_g\) \\
1.6 & DeepSeek-1.5B & - & - & \(m_s\) \\
1.7 & DeepSeek-1.5B & GIN & - & \(m_s + m_g\) \\
1.8 & DeepSeek-1.5B & Uni-mol & - & \(m_s + m_c\) \\
1.9 & DeepSeek-7B & 3D-GeoFormer & Highway & \(m_s + m_c\) \\
1.10 & DeepSeek-14B & Uni-mol & - & \(m_s + m_c\) \\
1.11 & DeepSeek-14B & GAT & Self Attention & \(m_s + m_g\) \\
1.12 & LLaMA-1B & - & - & \(m_s\) \\
1.13 & LLaMA-1B & CoATGIN & - & \(m_s + m_g\) \\
1.14 & LLaMA-1B & EGNN & Gated Fusion & \(m_s + m_c\) \\
1.15 & LLaMA-13B & SchNet & Bilinear Attention & \(m_s + m_c\) \\
\bottomrule
\end{tabular}
}
\vspace{0.1cm}
\begin{tabular}{p{0.9\textwidth}}
\textbf{Note:} \(m_s\) = molecular sequence, \(m_g\) = molecular graph, \(m_c\) = molecular conformation. '-' indicates that no method is applied.
\end{tabular}
\end{table*}

Given the flexibility of ModuLM, which enables a large number of potential model combinations, the goal of this section is not to exhaustively explore the entire model space. Instead, we select several model combinations as examples to demonstrate ModuLM’s robust capabilities in constructing and evaluating diverse model architectures across various datasets and performance metrics. It is worth noting that for different backbones, we adopt a unified pretraining strategy. The experimental backbones presented in the main text are LLMs pretrained based on the Structure Similarity-guided Grouping pretraining approach. 

During the evaluation phase on downstream tasks, we utilized the same datasets used in the MolTC framework \cite{fang2024moltc} for evaluating MRL tasks, including DrugBank (Version 5.0.3), ZhangDDI \cite{zhang2017predicting}, ChChMiner \cite{zitnik2018stanford}, DeepDDI \cite{ryu2018deep}, TWOSIDES \cite{tatonetti2012data}, Chromophore \cite{joung2020experimental}, MNSol \cite{marenich2020minnesota}, CompSol \cite{moine2017estimation}, Abraham \cite{grubbs2010mathematical}, CombiSolv \cite{vermeire2021transfer}, FreeSolv \cite{mobley2014freesolv}, and CombiSolv-QM \cite{vermeire2021transfer}. We further process these datasets using RDKit by generating multiple conformations for each molecule, aiming to explore the performance of ModuLM on 3D conformation-based LLM-driven MRL tasks. Based on ModuLM, we construct 15 models, as detailed in Table~\ref{tab:experimental_settings}, and compare them with five state-of-the-art LLM-based models for MRL tasks: Galactica, ChemT5 \cite{chemT5}, MolT5, MolCA \cite{MolCA}, and MolTC \cite{fang2024moltc}. 

For the DDI, SSI and CSI datasets, we randomly split the data into training, validation, and testing sets in a ratio of 7:2:1. Each experiment was repeated five times to mitigate the effects of randomness, and the average results were reported. All experiments were conducted using eight NVIDIA A100 80G GPUs. For more specific training details, please refer to the Appendix~\ref{app:exp_details}.

\subsection{Experimental Results and Analysis}

% \end{tabular}

% % \vspace{-0.8em}
% \end{table*}

\begin{table*}[ht]
\caption{Performance on DDI Datasets}
\label{tab:DDI_result}

\resizebox{\textwidth}{!}{
\begin{tabular}{
  c
  >{\centering\arraybackslash}p{2.3cm}
  >{\centering\arraybackslash}p{2.3cm}
  >{\centering\arraybackslash}p{2.3cm}
  >{\centering\arraybackslash}p{2.3cm}
  >{\centering\arraybackslash}p{2.3cm}
  >{\centering\arraybackslash}p{2.3cm}
}
\toprule
\textbf{Experiment} & \textbf{AUC-ROC (ChChMiner)} & \textbf{Accuracy (ChChMiner)} & \textbf{AUC-ROC (ZhangDDI)} & \textbf{Accuracy (ZhangDDI)} & \textbf{AUC-ROC (DeepDDI)} & \textbf{Accuracy (DeepDDI)}\\
\midrule
Chem T5\cite{christofidellis2023unifying} & 0.867 $\pm$ 0.012 & 0.814 $\pm$ 0.009 & 0.889 $\pm$ 0.017 & 0.751 $\pm$ 0.021 & 0.856 $\pm$ 0.012 & 0.784 $\pm$ 0.013 \\
MolCA\cite{MolCA} & 0.924 $\pm$ 0.006 & 0.901 $\pm$ 0.009 & 0.895 $\pm$ 0.006 & 0.745 $\pm$ 0.010 & 0.878 $\pm$ 0.014 & 0.841 $\pm$ 0.015 \\
MolT5\cite{edwards2022translation} & 0.914 $\pm$ 0.019 & 0.862 $\pm$ 0.022 & 0.901 $\pm$ 0.011 & 0.802 $\pm$ 0.015  & 0.907 $\pm$ 0.014 & 0.870 $\pm$ 0.016\\
MolTC\cite{fang2024moltc} & 0.964 $\pm$ 0.008 & 0.957 $\pm$ 0.006 & \textbf{0.941 $\pm$ 0.006} & 0.896 $\pm$ 0.008 & \textbf{0.977 $\pm$ 0.013} & 0.956 $\pm$ 0.011\\
\cmidrule(lr){1-7}
1.1 & 0.933 $\pm$ 0.011 & 0.924 $\pm$ 0.009 & 0.912 $\pm$ 0.008 & 0.854 $\pm$ 0.004 & 0.899 $\pm$ 0.010 & 0.855 $\pm$ 0.009\\
1.2 & 0.956 $\pm$ 0.008 & 0.943 $\pm$ 0.009 & 0.930 $\pm$ 0.006 & 0.872 $\pm$ 0.009 & 0.924 $\pm$ 0.008 & 0.887 $\pm$ 0.008\\
1.3 & 0.960 $\pm$ 0.010 & 0.954 $\pm$ 0.006 & 0.933 $\pm$ 0.007 & 0.891 $\pm$ 0.004 & 0.939 $\pm$ 0.007 & 0.904 $\pm$ 0.008\\
1.4 & 0.955 $\pm$ 0.005 & 0.949 $\pm$ 0.008 & 0.936 $\pm$ 0.014 & 0.901 $\pm$ 0.010 & 0.956 $\pm$ 0.008 & 0.919 $\pm$ 0.007\\
1.5 & 0.940 $\pm$ 0.009 & 0.932 $\pm$ 0.008 & 0.921 $\pm$ 0.008 & 0.866 $\pm$ 0.005 & 0.948 $\pm$ 0.009 & 0.912 $\pm$ 0.006 \\
1.6 & 0.936 $\pm$ 0.010 & 0.930 $\pm$ 0.011 & 0.920 $\pm$ 0.009 & 0.860 $\pm$ 0.008 & 0.906 $\pm$ 0.008 & 0.872 $\pm$ 0.008\\
1.7 & 0.957 $\pm$ 0.008 & 0.953 $\pm$ 0.010 & 0.934 $\pm$ 0.006 & 0.889 $\pm$ 0.004 & 0.958 $\pm$ 0.007 & 0.942 $\pm$ 0.007\\
1.8 & \textbf{0.966 $\pm$ 0.007} & \textbf{0.964 $\pm$ 0.005} & 0.938 $\pm$ 0.005 & \textbf{0.907 $\pm$ 0.006} & 0.972 $\pm$ 0.009 & \textbf{0.959 $\pm$ 0.010}\\
1.9 & 0.944 $\pm$ 0.010 & 0.935 $\pm$ 0.009 & 0.925 $\pm$ 0.005 & 0.870 $\pm$ 0.003 & 0.955 $\pm$ 0.008 & 0.930 $\pm$ 0.007\\
1.10 & 0.931 $\pm$ 0.012 & 0.918 $\pm$ 0.010 & 0.916 $\pm$ 0.009 & 0.861 $\pm$ 0.011 & 0.943 $\pm$ 0.010 & 0.915 $\pm$ 0.008\\
1.11 & 0.935 $\pm$ 0.007 & 0.921 $\pm$ 0.012 & 0.906 $\pm$ 0.009 & 0.855 $\pm$ 0.008 & 0.936 $\pm$ 0.009 & 0.908 $\pm$ 0.007\\
1.12 & 0.925 $\pm$ 0.013 & 0.911 $\pm$ 0.011 & 0.901 $\pm$ 0.008 & 0.852 $\pm$ 0.006 & 0.895 $\pm$ 0.010 & 0.850 $\pm$ 0.009\\
1.13 & 0.945 $\pm$ 0.009 & 0.937 $\pm$ 0.008 & 0.925 $\pm$ 0.007 & 0.870 $\pm$ 0.006 & 0.935 $\pm$ 0.008 & 0.904 $\pm$ 0.008\\
1.14 & 0.951 $\pm$ 0.007 & 0.946 $\pm$ 0.011 & 0.928 $\pm$ 0.004 & 0.875 $\pm$ 0.005 & 0.946 $\pm$ 0.007 & 0.918 $\pm$ 0.006\\
1.15 & 0.915 $\pm$ 0.016 & 0.896 $\pm$ 0.013 & 0.913 $\pm$ 0.008 & 0.860 $\pm$ 0.002 & 0.928 $\pm$ 0.011 & 0.897 $\pm$ 0.008\\
\bottomrule
\end{tabular}
}
\end{table*}

\vspace{-1em}
\begin{table*}[ht]
\caption{Performance on SSI Datasets}
\label{tab:SSI_result}
\resizebox{\textwidth}{!}{
\begin{tabular}{
  c
  >{\centering\arraybackslash}p{2.3cm}
  >{\centering\arraybackslash}p{2.3cm}
  >{\centering\arraybackslash}p{2.3cm}
  >{\centering\arraybackslash}p{2.3cm}
  >{\centering\arraybackslash}p{2.3cm}
  >{\centering\arraybackslash}p{2.3cm}
}
\toprule
\textbf{Experiment} & \textbf{MAE (FreeSolv)} & \textbf{RMSE (FreeSolv)} & \textbf{MAE (CompSol)} & \textbf{RMSE (CompSol)} & \textbf{MAE (CombiSolv)} & \textbf{RMSE (CombiSolv)}\\
\midrule
Chem T5\cite{christofidellis2023unifying} & 0.923 $\pm$ 0.022 & 1.511 $\pm$ 0.043 & 0.611 $\pm$ 0.017 & 0.766 $\pm$ 0.032 & 0.840 $\pm$ 0.040 & 1.294 $\pm$ 0.043 \\
MolCA\cite{MolCA} & 0.761 $\pm$ 0.034 & 1.303 $\pm$ 0.039 & 0.505 $\pm$ 0.036 & 0.726 $\pm$ 0.040 & 0.771 $\pm$ 0.033 & 1.130 $\pm$ 0.027 \\
MolT5\cite{edwards2022translation} & 0.733 $\pm$ 0.047 & 1.135 $\pm$ 0.059 & 0.496 $\pm$ 0.028 & 0.708 $\pm$ 0.020  & 0.677 $\pm$ 0.024 & 1.066 $\pm$ 0.027\\
MolTC\cite{fang2024moltc} & 0.533 $\pm$ 0.018 & 0.726 $\pm$ 0.022 & 0.244 $\pm$ 0.018 & 0.356 $\pm$ 0.022 & 0.237 $\pm$ 0.019 & 0.465 $\pm$ 0.022\\
\cmidrule(lr){1-7}
1.1 & 0.710 $\pm$ 0.021 & 1.120 $\pm$ 0.030 & 0.472 $\pm$ 0.024 & 0.665 $\pm$ 0.028 & 0.615 $\pm$ 0.026 & 0.984 $\pm$ 0.032 \\
1.2 & 0.570 $\pm$ 0.020 & 0.910 $\pm$ 0.028 & 0.384 $\pm$ 0.022 & 0.540 $\pm$ 0.025 & 0.568 $\pm$ 0.023 & 0.930 $\pm$ 0.030 \\
1.3 & 0.556 $\pm$ 0.018 & 0.840 $\pm$ 0.025 & 0.366 $\pm$ 0.021 & 0.522 $\pm$ 0.024 & 0.487 $\pm$ 0.021 & 0.820 $\pm$ 0.027 \\
1.4 & 0.534 $\pm$ 0.017 & 0.808 $\pm$ 0.024 & 0.347 $\pm$ 0.020 & 0.501 $\pm$ 0.023 & 0.447 $\pm$ 0.020 & 0.780 $\pm$ 0.026 \\
1.5 & 0.580 $\pm$ 0.016 & 0.972 $\pm$ 0.023 & 0.403 $\pm$ 0.019 & 0.575 $\pm$ 0.021 & 0.579 $\pm$ 0.019 & 0.898 $\pm$ 0.025 \\
1.6 & 0.685 $\pm$ 0.015 & 1.086 $\pm$ 0.022 & 0.451 $\pm$ 0.018 & 0.643 $\pm$ 0.020 & 0.602 $\pm$ 0.018 & 0.945 $\pm$ 0.024 \\
1.7 & 0.550 $\pm$ 0.014 & 0.749 $\pm$ 0.021 & 0.271 $\pm$ 0.017 & 0.415 $\pm$ 0.019 & 0.289 $\pm$ 0.017 & 0.515 $\pm$ 0.023 \\
1.8 & \textbf{0.510 $\pm$ 0.013} & \textbf{0.698 $\pm$ 0.020} & \textbf{0.191 $\pm$ 0.016} & \textbf{0.298 $\pm$ 0.018} & \textbf{0.190 $\pm$ 0.016} & \textbf{0.388 $\pm$ 0.022} \\
1.9 & 0.555 $\pm$ 0.014 & 0.825 $\pm$ 0.021 & 0.335 $\pm$ 0.017 & 0.490 $\pm$ 0.019 & 0.393 $\pm$ 0.017 & 0.595 $\pm$ 0.023 \\
1.10 & 0.605 $\pm$ 0.015 & 0.850 $\pm$ 0.022 & 0.443 $\pm$ 0.018 & 0.598 $\pm$ 0.020 & 0.548 $\pm$ 0.018 & 0.864 $\pm$ 0.024 \\
1.11 & 0.590 $\pm$ 0.016 & 0.876 $\pm$ 0.023 & 0.458 $\pm$ 0.019 & 0.601 $\pm$ 0.021 & 0.556 $\pm$ 0.019 & 0.839 $\pm$ 0.025 \\
1.12 & 0.745 $\pm$ 0.017 & 1.091 $\pm$ 0.024 & 0.514 $\pm$ 0.020 & 0.692 $\pm$ 0.022 & 0.687 $\pm$ 0.020 & 1.008 $\pm$ 0.026 \\
1.13 & 0.605 $\pm$ 0.016 & 0.880 $\pm$ 0.023 & 0.374 $\pm$ 0.019 & 0.530 $\pm$ 0.021 & 0.538 $\pm$ 0.019 & 0.820 $\pm$ 0.025 \\
1.14 & 0.580 $\pm$ 0.015 & 0.887 $\pm$ 0.022 & 0.360 $\pm$ 0.018 & 0.515 $\pm$ 0.020 & 0.460 $\pm$ 0.018 & 0.790 $\pm$ 0.024 \\
1.15 & 0.630 $\pm$ 0.018 & 0.910 $\pm$ 0.025 & 0.450 $\pm$ 0.021 & 0.633 $\pm$ 0.023 & 0.567 $\pm$ 0.021 & 0.892 $\pm$ 0.027\\
\bottomrule
\end{tabular}
}
\end{table*}

Table~\ref{tab:DDI_result} and Table~\ref{tab:SSI_result} show the experimental results of commonly used MRL LLMs, as well as their performance under our experimental settings. It is worth noting that in our experimental setup, we adopt a direct inference approach using LLMs without employing chain-of-thought reasoning. 
In our setup, we design models with different backbones, input formats, and encoders. Due to space limitations, additional experimental results are presented in the Appendix~\ref{app:more_exp}.

\textbf{Impact of Input Data and Encoders:} We use the settings \{1.1, 1.2, 1.4\}, \{1.6, 1.7, 1.8\}, and \{1.12, 1.13\} to evaluate ModuLM's ability to analyze and integrate different inputs and encoder layers within LLMs. The results in the table indicate that integrating multimodal information, such as 2D molecular graphs and 3D molecular conformations, can indeed enhance model performance. Notably, models that incorporate 3D molecular conformation information achieve the best results.

\textbf{Impact of Interaction Layers:} We use the configurations \{1.3, 1.5, 1.9, 1.11, 1.14, 1.15\} to evaluate ModuLM’s ability to analyze the effect of feature interaction layers. We first conduct experiments with various non-interaction designs. The analysis shows that adding interaction layers consistently improves model performance to some extent. This confirms the importance of interaction layers in LLM-based MRL models. However, existing LLM comparisons generally ignore multi-molecular interaction information. ModuLM enables multi-dimensional analysis, which helps better assess the impact of different types of multimodal information on model performance.

\textbf{Impact of Different Backbones:} Existing LLM-based MRL tasks generally lack systematic evaluation across different backbones. However, thoroughly testing LLMs under various experimental configurations requires significant time and resources. ModuLM offers a framework for efficient and rapid evaluation. As shown in Table~\ref{tab:DDI_result} and Table ~\ref{tab:SSI_result}, we conduct experiments using different backbones. Among them, models from the DeepSeek series often achieve better performance. Interestingly, our results reveal that larger model sizes do not necessarily lead to better performance in MRL tasks. This may be because larger LLMs possess stronger generalization ability, which can limit task-specific adaptation during fine-tuning. In contrast, smaller LLMs adapt better during fine-tuning, leading to stronger task specialization in MRL scenarios.

\subsection{Custom Model Design and Evaluation}
This section demonstrates how users can leverage ModuLM to extend, construct, and analyze more complex models.
\begin{figure}[ht]  % 可以指定浮动的位置，例如[h]表示在当前位置
    \centering  % 图片居中显示
    \includegraphics[width=1\textwidth]{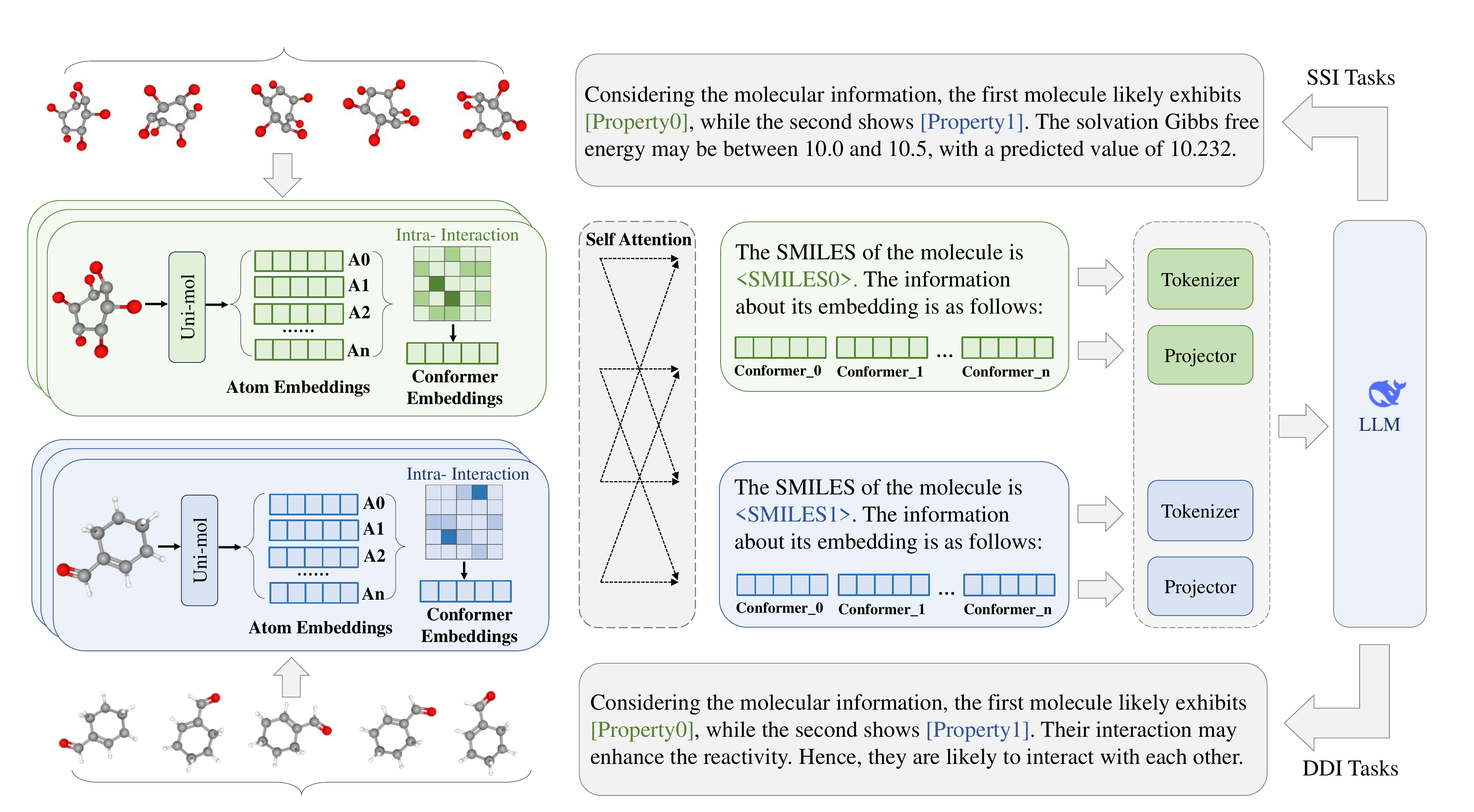}  % 这里设置图片的路径及尺寸
    \caption{An overview of the custom model designed using ModuLM.}  % 添加标题
    \label{fig:custom}  % 可选：为图片设置标签，方便在文中引用
\end{figure}
Figure~\ref{fig:custom} presents an overview of the newly proposed model architecture. This design builds upon the best-performing configuration identified in Experiment 1.8, which utilizes 3D molecular conformations, employs Uni-Mol as the encoder, and adopts DeepSeek-1.5B as the LLM backbone. In this enhanced version, we further refined the encoder and the dual-molecule interaction mechanism, and incorporated CoT prompting strategy to boost the model's reasoning capabilities.

To simulate practical scenarios in which users may wish to combine predefined encoders with custom modules, we reimplement the molecular encoder as a user-defined component and integrate it with other existing encoder modules to construct a complete, functional model. Specifically, we redesign the representation of atom-level data encoded by Uni-Mol by calculating the attention weights of each atom to all other atoms within the conformation. These weights are then used to reaggregate the atom features, and the mean of the weighted features serves as the new conformation-level representation. The full model construction process based on ModuLM is detailed in Appendix~\ref{app:use}.

We retained the experimental settings described in Section~\ref{wholeexp} and conducted ablation studies to reflect real-world use cases of ModuLM. In these experiments, \textit{w/o M-Encoder} refers to the exclusion of the customized encoder, \textit{w/o Interaction} indicates the removal of the molecule interaction design, and \textit{w/o CoT} represents the absence of the chain-of-thought reasoning mechanism. The full experimental results are reported in Table~\ref{tab:mresult}, while the outcomes of the ablation studies are summarized in Table~\ref{tab:aresult}.

\vspace{-1em}
\begin{table}[!htbp]
\caption{Performance of Custom Model on DDI and SSI Datasets}
\label{tab:mresult}

\resizebox{\textwidth}{!}{
\begin{tabular}{
  c
  >{\centering\arraybackslash}p{2.3cm}
  >{\centering\arraybackslash}p{2.3cm}
  >{\centering\arraybackslash}p{2.3cm}
  >{\centering\arraybackslash}p{2.3cm}
  >{\centering\arraybackslash}p{2.3cm}
  >{\centering\arraybackslash}p{2.3cm}
}
\toprule
\textbf{Experiment} & \textbf{Accuracy (ChChMiner)} & \textbf{Accuracy (ZhangDDI)} & \textbf{Accuracy (DeepDDI)} & \textbf{RMSE (FreeSolv)} & \textbf{RMSE (CompSol)} & \textbf{RMSE (CombiSolv)}\\
\midrule
1.7 & 0.953 $\pm$ 0.010 & 0.889 $\pm$ 0.004 & 0.942 $\pm$ 0.007 & 0.749 $\pm$ 0.021 & 0.415 $\pm$ 0.019 & 0.515 $\pm$ 0.023 \\
1.8 & 0.964 $\pm$ 0.005 & 0.907 $\pm$ 0.006 & 0.959 $\pm$ 0.010 & 0.698 $\pm$ 0.020 & 0.298 $\pm$ 0.018 & 0.388 $\pm$ 0.022 \\
Custom Model & \textbf{0.968 $\pm$ 0.006} & \textbf{0.911 $\pm$ 0.006} & \textbf{0.964 $\pm$ 0.008} & \textbf{0.680 $\pm$ 0.019} & \textbf{0.288 $\pm$ 0.013} & \textbf{0.359 $\pm$ 0.013} \\
\bottomrule
\end{tabular}
}
\end{table}

\vspace{-1em}
\begin{table}[ht]
\caption{Results of the Ablation Study on DDI and SSI Datasets}
\label{tab:aresult}
\resizebox{\textwidth}{!}{
\begin{tabular}{
  c
  >{\centering\arraybackslash}p{2.3cm}
  >{\centering\arraybackslash}p{2.3cm}
  >{\centering\arraybackslash}p{2.3cm}
  >{\centering\arraybackslash}p{2.3cm}
  >{\centering\arraybackslash}p{2.3cm}
  >{\centering\arraybackslash}p{2.3cm}
}
\toprule
\textbf{Experiment} & \textbf{Accuracy (ChChMiner)} & \textbf{Accuracy (ZhangDDI)} & \textbf{Accuracy (DeepDDI)} & \textbf{RMSE (FreeSolv)} & \textbf{RMSE (CompSol)} & \textbf{RMSE (CombiSolv)}\\
\midrule
w/o M-Encoder   & 0.962 $\pm$ 0.007 & 0.905 $\pm$ 0.005 & 0.959 $\pm$ 0.007 & 0.700 $\pm$ 0.019 & 0.299 $\pm$ 0.017 & 0.370 $\pm$ 0.020 \\
w/o Interaction & 0.955 $\pm$ 0.008 & 0.896 $\pm$ 0.006 & 0.950 $\pm$ 0.007 & 0.705 $\pm$ 0.018 & 0.317 $\pm$ 0.016 & 0.388 $\pm$ 0.019 \\
w/o CoT         & 0.959 $\pm$ 0.006 & 0.901 $\pm$ 0.006 & 0.956 $\pm$ 0.008 & 0.732 $\pm$ 0.020 & 0.335 $\pm$ 0.018 & 0.382 $\pm$ 0.021 \\
Full Model      & \textbf{0.968 $\pm$ 0.006} & \textbf{0.911 $\pm$ 0.006} & \textbf{0.964 $\pm$ 0.008} & \textbf{0.680 $\pm$ 0.019} & \textbf{0.288 $\pm$ 0.013} & \textbf{0.359 $\pm$ 0.013} \\
\bottomrule
\end{tabular}
}
\end{table}

Through the above experiments, we validated ModuLM’s strong capability in supporting user-defined models. Users can define custom encoders following our provided protocol and flexibly integrate them with other encoders and interaction layers as modular components. The customized model involves a more complex configuration, including user-defined blocks and additional operations such as stacking and flattening the outputs of ModuLM components. Moreover, with ModuLM’s dynamic model construction mechanism, users can easily adjust the model structure to perform ablation studies. As shown in Table~\ref{tab:mresult} both the CoT reasoning prompt and the interaction layer contribute to performance improvements. Additionally, our custom encoder module, which incorporates internal interaction mechanisms, also enhances model performance.

\section{Conclusion}
We propose ModuLM, a framework that supports the dynamic construction of LLM-based MRL models to address benchmarking challenges in LLM-driven molecular relational learning. ModuLM accommodates multiple molecular input formats and enables the flexible assembly of diverse model architectures, facilitating robust and scalable experimentation. The framework simplifies model development and standardizes the evaluation process across different architectures, ensuring fair and consistent benchmarking. By providing a flexible and modular platform, ModuLM not only advances research in the MRL field but also lays the foundation for cross-disciplinary collaboration and innovation, with broad applications in areas such as drug design and molecular interaction analysis, helping researchers accelerate critical biomedical discoveries.

\textbf{Limitations:} The benchmark experiments presented in this paper demonstrate the capabilities of the ModuLM framework in constructing and comparing various model architectures, yet they do not cover all possible model combinations. Our primary goal is to highlight ModuLM's advantages in enabling flexible model construction and efficient model comparison, showcasing its adaptability and effectiveness across diverse model architectures. While the current experiments provide valuable insights into the functionality and potential of the framework, a comprehensive exploration of all possible model combinations is beyond the scope of this study and is intended as a key direction for future research. We encourage the research community to further investigate the performance of different configurations using our framework, thus promoting the diversity and innovation of model architectures. 

\textbf{Future Work:} In the future, we will continue to maintain and expand the components within the ModuLM framework to further enhance its flexibility and applicability. Specifically, we will focus on introducing additional types of encoders, interaction layers, and evaluation metrics, while also expanding support for a broader range of LLMs to improve the generalizability and scalability of the framework. At the same time, we will deepen ModuLM’s application in molecular relational learning tasks, particularly in areas such as drug discovery and protein interaction analysis. We expect that through continuous iteration and optimization, ModuLM will become a powerful tool for advancing molecular relational learning and interdisciplinary research, providing flexible and efficient technical support for future scientific endeavors.

% \section*{References}
\bibliographystyle{plain} 
\bibliography{anthology,custom,mybenchmark,neurips_data_2024,reference} 

% References follow the acknowledgments in the camera-ready paper. Use unnumbered first-level heading for
% the references. Any choice of citation style is acceptable as long as you are
% consistent. It is permissible to reduce the font size to \verb+small+ (9 point)
% when listing the references.
% Note that the Reference section does not count towards the page limit.
% \medskip

% {
% \small

% [1] Alexander, J.A.\ \& Mozer, M.C.\ (1995) Template-based algorithms for
% connectionist rule extraction. In G.\ Tesauro, D.S.\ Touretzky and T.K.\ Leen
% (eds.), {\it Advances in Neural Information Processing Systems 7},
% pp.\ 609--616. Cambridge, MA: MIT Press.

% [2] Bower, J.M.\ \& Beeman, D.\ (1995) {\it The Book of GENESIS: Exploring
%   Realistic Neural Models with the GEneral NEural SImulation System.}  New York:
% TELOS/Springer--Verlag.

% [3] Hasselmo, M.E., Schnell, E.\ \& Barkai, E.\ (1995) Dynamics of learning and
% recall at excitatory recurrent synapses and cholinergic modulation in rat
% hippocampal region CA3. {\it Journal of Neuroscience} {\bf 15}(7):5249-5262.
% }

%%%%%%%%%%%%%%%%%%%%%%%%%%%%%%%%%%%%%%%%%%%%%%%%%%%%%%%%%%%%

%%%%%%%%%%%%%%%%%%%%%%%%%%%%%%%%%%%%%%%%%%%%%%%%%%%%%%%%%%%%

% \bibliographystyle{plain}

\newpage
\appendix
\section{Appendix}
\subsection{General Information}
\subsubsection{Links}
The code for ModuLM is currently available in our code repository~\url{https://anonymous.4open.science/r/ModuLM}. The repository provides reproduction scripts for the best-performing model, and if you wish to reproduce other experiments conducted in this paper, please refer to the Appendix~\ref{app:use}.
\subsubsection{Licenses}
The ModuLM  is under the CC BY 4.0. We, the authors, bear all responsibility in case of
 violation of rights.

\subsection{Details of Datasets}
\label{app:datasets}
In ModuLM, we provide diverse datasets for DDI, SSI, and CSI tasks to comprehensively evaluate ModuLM's  flexibility and generalizability. An overview of the datasets is provided in Table~\ref{tab:data_stats}.

\begin{table}[!htbp]\footnotesize
% \small
    \caption{Data statistics.}
    \label{tab:data_stats}
    \vspace{1.0em}
    \centering
    % \vspace{-0.6em}
    \setlength{\tabcolsep}{9.0pt}
    \def\arraystretch{1.05}

    \begin{tabular}{c|cc|ccc}
    \toprule
    \multicolumn{1}{c|}{Task} & \multicolumn{2}{c|}{Dataset} & \ $\mathcal{G}^{1}$ & \ $\mathcal{G}^{2}$ & \ Pairs \\ \hline 
    \multirow{5}{*}{DDI} 
    % & \multicolumn{2}{c|}{DrugBank} & 1704 & 1704 & 191,400 \\
    & \multicolumn{2}{c|}{ZhangDDI} & 548 & 548 & 48,548 \\
    & \multicolumn{2}{c|}{ChChMiner} & 1322 & 1322 & 48,514 \\
    & \multicolumn{2}{c|}{DeepDDI} & - & - & 192,284 \\
    & \multicolumn{2}{c|}{TWOSIDES} & 555 & 555 & 3,576,513 \\ \hline
    \multirow{6}{*}{SSI} 
    & \multicolumn{2}{c|}{MNSol} & 372 & 86 & 2,275 \\
    & \multicolumn{2}{c|}{FreeSolv} & 560 & 1 & 560 \\
    & \multicolumn{2}{c|}{CompSol} & 442 & 259 & 3,548 \\
    & \multicolumn{2}{c|}{Abraham} & 1038 & 122 & 6,091 \\
    & \multicolumn{2}{c|}{CombiSolv} & 1,368 & 291 & 10,145 \\
    & \multicolumn{2}{c|}{CombiSolv-QM} & 11,029 & 284 & 1,000,000 \\ \hline
    \multirow{1}{*}{CSI} 
    & \multicolumn{2}{c|}{Chromophore} & 7,016 & 365 & 20,236 \\ 
    \bottomrule
    \end{tabular}

\end{table}

\vspace{3pt}
\noindent\textbf{ZhangDDI}~\cite{zhang2017predicting}.
Includes 548 drugs and 48,548 labeled interactions, incorporating various drug similarity metrics for similarity-aware DDI modeling. It is commonly used for evaluating similarity-based prediction methods.

\noindent\textbf{ChChMiner}~\cite{zitnik2018stanford}.
Provides 1,322 drugs and 48,514 interactions sourced from FDA labels and literature, emphasizing clinically verified DDIs. It is suitable for real-world DDI risk assessment.

\noindent\textbf{DeepDDI}~\cite{ryu2018deep}.
Contains 192,284 labeled DDIs with side effect annotations extracted from DrugBank. The dataset supports multi-label classification tasks involving adverse interactions.

\noindent\textbf{TWOSIDES}~\cite{tatonetti2012data}.
Offers 555 drugs and over 3.5 million interactions with 1,318 types, derived from adverse event reports. It enables exploration of long-tail DDI patterns at scale.

\noindent\textbf{Chromophore}~\cite{joung2020experimental}.
Includes 20,236 chromophore–solvent pairs with optical properties like absorption, emission, and lifetime. Unreliable entries are removed, and log-normalization is applied to reduce target skew.

\noindent\textbf{MNSol}~\cite{marenich2020minnesota}.
Provides 3,037 solvation or transfer free energy records across 790 solutes and 92 solvents. We use 2,275 filtered combinations consistent with prior benchmarks.

\noindent\textbf{FreeSolv}~\cite{mobley2014freesolv}.
Offers 643 hydration free energy measurements in water for small molecules. We retain 560 experimental entries for aqueous solvation prediction tasks.

\noindent\textbf{CompSol}~\cite{moine2017estimation}.
Contains 3,548 solute–solvent pairs focusing on hydrogen-bonding effects in solvation energy. It evaluates the impact of molecular interactions on solvent behavior.

\noindent\textbf{Abraham}~\cite{grubbs2010mathematical}.
Covers 6,091 combinations from 1,038 solutes and 122 solvents based on Abraham's solvation parameter model. It supports solvent effect modeling via linear energy relationships.

\noindent\textbf{CombiSolv}~\cite{vermeire2021transfer}.
Combines MNSol, FreeSolv, CompSol, and Abraham into a unified dataset of 10,145 solute–solvent combinations. It serves as a comprehensive benchmark for solvation prediction.

\noindent\textbf{CombiSolv-QM}~\cite{vermeire2021transfer}.
Includes 1 million QM-generated solute–solvent pairs with diverse chemical compositions. It evaluates large-scale generalization and robustness of molecular models.

\subsection{Details of ModuLM Components
 }
 \label{app:components}
 \subsubsection{2D Graph Encoders}
\noindent\textbf{GCN}~\cite{kipf2017semi}. Performs graph convolution using a normalized Laplacian to aggregate neighboring node features, enabling semi-supervised learning on graph-structured data.

\noindent\textbf{MPNN}~\cite{gilmer2017neural}. Generalizes message-passing in graphs by learning both message functions and update rules, providing a flexible framework for molecular property prediction.

\noindent\textbf{GAT}~\cite{velickovic2018graph}. Applies self-attention to adaptively weigh neighbor contributions during node embedding updates, improving learning on irregular graph structures.

\noindent\textbf{NeuralFP}~\cite{duvenaud2015convolutional}. Learns molecular fingerprints via graph convolutional layers, enabling end-to-end learning of molecular representations for property prediction.

\noindent\textbf{AttentiveFP}~\cite{Xiong_Wang_2019b}. Utilizes attention mechanisms to prioritize significant molecular substructures, enhancing interpretability and predictive performance in QSAR tasks.

\noindent\textbf{GIN}~\cite{xu2019how}. Enhances graph discriminative power with injective neighborhood aggregation, theoretically achieving maximal expressive capacity among GNNs.

\noindent\textbf{GraphSAGE}~\cite{hamilton2017inductive}. Generates embeddings by sampling and aggregating neighbor features in an inductive manner, making it scalable to large dynamic graphs.

\noindent\textbf{CoATGIN}~\cite{zhang2022coatgin}. Integrates convolutional aggregation and attention mechanisms to capture both local motifs and global contexts in molecular graphs.

\subsubsection{3D Conformation Encoders}

\noindent\textbf{EGNN}~\cite{satorras2021n}. Preserves equivariance to Euclidean transformations by jointly updating node features and coordinates, enabling effective modeling of molecular geometry.

\noindent\textbf{3D-GeoFormer}~\cite{zhou2023self}. Leverages geometric transformers to model spatial relations and interactions in molecular structures, enhancing 3D representation learning with attention.

\noindent\textbf{SE(3)-Transformer}~\cite{fuchs2020se}. Incorporates SE(3) equivariance using tensor field networks, making it well-suited for complex 3D molecular and protein data.

\noindent\textbf{PaiNN}~\cite{schutt2021equivariant}. Achieves rotational equivariance by separating scalar and vector features in message passing, allowing accurate force field and energy predictions.

\noindent\textbf{GVP}~\cite{jing2021learning}. Combines geometric vector perceptrons with scalar and vector features for molecular structure modeling, while being applicable to general 3D molecular graphs.

\noindent\textbf{GearNet}~\cite{zhang2022protein}. Builds multi-scale molecular representations by integrating structural and sequential information, enhancing learning across different biological levels.

\noindent\textbf{DimeNet++}~\cite{gasteiger2020directional}. Captures directional information and angular relations in molecular graphs with improved message passing and higher efficiency than its predecessor.

\noindent\textbf{SchNet}~\cite{schutt2017schnet}. Models quantum interactions using continuous-filter convolutions, enabling accurate predictions of atomic-level properties in molecular systems.

\noindent\textbf{SphereNet}~\cite{liu2022spherical}. Encodes spherical angles and radial distances to capture 3D molecular geometry more precisely, leading to improved performance in quantum property prediction.

\noindent\textbf{G-SphereNet}~\cite{luo2022autoregressive}. Extends SphereNet with an autoregressive generative mechanism, enabling the modeling and generation of complex molecular conformations.

\noindent\textbf{Uni-Mol}~\cite{zhou2023uni}. Unifies molecular pretraining and finetuning in 3D space using a transformer-based architecture, supporting diverse downstream tasks with spatial awareness.

\subsubsection{Interaction Layers}
\noindent \textbf{Bilinear Attention}~\cite{Bai_Miljković_John_Lu_2023}. The Bilinear Attention Network (BAN) layer captures interactions between 2D feature sets through bilinear transformations, followed by attention pooling and batch normalization.

\noindent \textbf{Self Attention}~\cite{vaswani2017attention}. Self-attention mechanisms allow a feature set to focus on its own elements, enabling models to capture relationships within the same source of data.

\noindent \textbf{Cross Attention}~\cite{Qian_Li_Wu_Zhang_2023}. Cross-attention captures interactions between two distinct feature sets by applying attention mechanisms that focus on cross-source dependencies.

\noindent \textbf{Highway}~\cite{Zhu_Zhao_Wen_Wang_Wang_2023}. The Highway mechanism combines 1D features through gated layers, allowing information to flow selectively by controlling the gates.

\noindent \textbf{Gated Fusion}~\cite{ren2018gated}. Gated fusion combines 1D features from two sources by applying gated transformations, producing a unified representation that captures the interactions between them.

\noindent \textbf{Bilinear Fusion}~\cite{lin2015bilinear}. Bilinear Fusion combines 1D features using a bilinear transformation and ReLU activation, capturing multiplicative interactions to enhance feature representation.

\noindent \textbf{Mean}. The Mean method combines feature sets by averaging their values.

\subsubsection{Backbones}
In ModuLM, we provide seven popular LLMs: DeepSeek-1.5B, DeepSeek-7B, DeepSeek-14B~\cite{bi2024deepseek}, LLaMA-1B, LLaMA-13B~\cite{touvron2023llama}, Galactica-1.3B, and Galactica-6.7B~\cite{taylor2022galactica}.

\noindent\textbf{DeepSeek.} The DeepSeek-1.5B, DeepSeek-7B, and DeepSeek-14B models are derived from the Qwen-2.5 series, which were originally licensed under the Apache 2.0 License, and have now been fine-tuned with 800k samples curated using DeepSeek-R1. 

\noindent\textbf{LLama.} LLama 1B incorporated logits from the LLama 3.1 8B and 70B models during the pretraining stage, using the outputs (logits) from these larger models as token-level targets. Knowledge distillation was applied after pruning to recover performance. LLama-13B was pretrained on 2 trillion tokens of data from publicly available sources and fine-tuned on publicly available instruction datasets, along with over one million new human-annotated examples, making it a general-purpose LLM.

\noindent\textbf{Galactica.} Galactica-1.3B and Galactica-6.7B are large language models developed by Meta for scientific research and knowledge-intensive tasks. These models are designed to assist with tasks in fields like scientific literature, research summarization, and computational biology.

\begin{table*}[!htbp]\centering
\begin{minipage}{\textwidth}\vspace{0mm}    \centering
\caption{Prompts settings.}
    \label{tab:prompt2}
\begin{tcolorbox} 
    \centering
   
      \footnotesize
    \begin{tabular}{p{0.97\textwidth} c}
   \textbf{ {\bf Direct  Reasoning for Qualitative Tasks} } &\\
The first molecule \emb{<SMILES0>} and the second molecule \newemb{<SMILES1>} are expected to interact with each other, potentially forming a molecular complex or influencing each other's properties.
\\
   \textbf{ {\bf CoT-based  Reasoning for Qualitative Tasks} } &\\
The first molecule \emb{<SMILES0>} is likely to exhibit \emb{[Property0]}, while the second molecule \newemb{<SMILES1>} is likely to exhibit \newemb{[Property1]}. Hence, the first drug molecule may alter the therapeutic effects of the second drug molecule. Therefore, they are likely to interact with each other.

% \textbf{ {\bf Target Response for DDI Tasks (Fine-tuning)} } & \\
% Considering the molecular conformation, the first molecule may exhibit the property \emb{[Property0]}, while the second molecule may possess the property \texttt{[Property1]}. This interaction could potentially enhance the photoreactivity of the second molecule. Therefore, they are likely to interact with each other.

    \hrulefill & \\
    
     \textbf{ {\bf Direct  Reasoning for Quantitative Tasks} } &\\
The solvation Gibbs free energy between the first molecule \emb{<SMILES0>} and the second molecule \newemb{<SMILES1>} is 4.6232.\\
      \textbf{ {\bf CoT-based  Reasoning for Quantitative Tasks} } &\\
The first molecule \emb{<SMILES0>} is likely to exhibit \emb{[Property0]}, while the second molecule \newemb{<SMILES1>} is likely to exhibit \newemb{[Property1]}. Therefore, their solvation Gibbs free energy is likely to fall between 4.0 and 4.5, with a precise value potentially being 4.6232.
    \end{tabular}
\end{tcolorbox}
\vspace{-2mm}

\end{minipage}
\end{table*}

\subsubsection{Prompts}
\label{app:prompt}
In the main text, we mentioned the two prompts involved in our experiments: Direct and Chain-of-Thought-based reasoning. The prompt design is shown in the Table~\ref{tab:prompt2}.

\subsection{Example Usage of ModuLM}
\label{app:use}
This section demonstrates how to construct example models from the main text using ModuLM. In ModuLM, we provide a configuration method based on a JSON file. In the modules we offer, users only need to modify the corresponding parameters. It is worth noting that some of the parameters below are included solely to demonstrate the comprehensiveness of our framework; if the goal is simply to use it, most hyperparameters do not need to be changed.

\subsubsection{Loading the Dataset}
Taking the DeepDDI dataset loading as an example, the path to the dataset is defined using the \texttt{root} parameter. To accelerate data loading, the \texttt{num\_workers} parameter is used to enable multi-threaded data processing. Additionally, the \texttt{use\_3d} flag controls whether to incorporate 3D molecular conformational data as input. This allows users to flexibly switch between 2D and 3D molecular representations depending on the task requirements and available structural information.

\begin{minted}[fontsize=\footnotesize]{json}
{
    "root": "data/DDI/DeepDDI/",
    "num_workers": 5,
    "use_3d":true
}
\end{minted}

\subsubsection{Initializing Encoder}
In this setup, we utilize the Uni-Mol model as our 3D molecular conformation encoder. The specific encoder is selected by setting the \texttt{graph3d} parameter accordingly. To fine-tune its behavior and architecture, we provide a dedicated configuration file that defines key hyperparameters such as the number of layers, embedding dimensions, attention mechanisms, and dropout rates. This modular design allows for flexible customization and seamless integration into various molecular representation learning tasks.

\begin{minted}[fontsize=\footnotesize]{json}
{
    "graph3d": "unimol",
    "con_activation_dropout": 0.0,
    "con_activation_fn": "gelu",
    "con_attention_dropout": 0.1,
    "con_delta_pair_repr_norm_loss": -1.0,
    "con_dropout": 0.1,
    "con_emb_dropout": 0.1,
    "con_encoder_attention_heads": 64,
    "con_encoder_embed_dim": 512,
    "con_encoder_ffn_embed_dim": 2048,
    "con_encoder_layers": 15,
    "con_max_atoms": 256,
    "con_max_seq_len": 512
}
\end{minted}
It is important to note that in order to build the example model provided in the main text, we need to make further modifications and extensions to the Uni-Mol model. Once the code has been extended, we can simply place it in the specified directory and continue managing and calling it through the config file.
\subsubsection{Configuring LLM
}
Here, the \texttt{mode} specifies the model's mode, whether it is pretraining or fine-tuning. The backbone parameter indicates the LLM to be used, while \texttt{min\_len} and \texttt{max\_len} define the minimum and maximum lengths of the generated text. Additional details and parameters for LLM-based text generation are provided in our code repository; this section highlights only a few key settings as examples.

\begin{minted}[fontsize=\footnotesize]{json}
{   "mode":"ft",
    "backbone": "DeepSeek-1.5B",
    "min_len": 10,
    "max_len": 40
}
\end{minted}

\subsubsection{Training the Model}
After constructing the model, we can fine-tune it by configuring the appropriate LoRA file. To make it easier for others to fine-tune using our framework, we provide the LoRA parameter configuration for model training here.
\begin{minted}[fontsize=\footnotesize]{json}
{
    "base_model_name_or_path": null,
    "bias": "none",
    "fan_in_fan_out": false,
    "inference_mode": false,
    "init_lora_weights": true,
    "lora_alpha": 32,
    "lora_dropout": 0.1,
    "target_modules": ["q_proj", "v_proj", "out_proj", "fc1", "fc2"],
    "peft_type": "LORA",
    "r": 16,
    "modules_to_save": null,
    "task_type": "CAUSAL_LM"
}
\end{minted}
The specific batch size and number of training epochs are also configured in a unified manner. Here, the \texttt{batch\_size} specifies the number of samples processed in each training step, which in this case is set to 12. The \texttt{max\_epochs} defines the total number of training iterations over the entire dataset, set here to 20 epochs. The \texttt{save\_every\_n\_epochs} parameter indicates that the model's state will be saved every 5 epochs. The \texttt{scheduler} field specifies the learning rate scheduling strategy—\texttt{linear\_warmup\_cosine\_lr} gradually increases the learning rate during a warm-up period, then decays it following a cosine curve. The \texttt{seed} ensures reproducibility of training results by fixing randomness. \texttt{warmup\_lr} and \texttt{warmup\_steps} define the initial learning rate and the number of steps over which it will warm up, respectively. Lastly, \texttt{weight\_decay} is used as a regularization technique to prevent overfitting by penalizing large weights during optimization.

\begin{minted}[fontsize=\footnotesize]{json}
{
    "batch_size": 12,
    "max_epochs": "30",
    "save_every_n_epochs": 5,
    "scheduler": linear_warmup_cosine_lr,
    "seed": 42,
    "warmup_lr": 1e-06,
    "warmup_steps": 1000,
    "weight_decay": "0.05"
}
\end{minted}

\subsection{More experimental details
}
\label{app:exp_details}

\subsubsection{Details of Experimental Setup}
In this section, we provide a more detailed description of the experimental data and testing configurations used in the main text.

\noindent \textbf{Training Epochs.}
At the beginning of each experiment, we initiate incremental pretraining by running 5 epochs on the collected pretraining dataset. During the subsequent fine-tuning stage, the number of training epochs is task-dependent. Specifically, for the DDI task, we fine-tune the model for 50 epochs. For SSI datasets containing more than 3000 molecular pairs, we adopt a two-stage fine-tuning strategy: first, the model is fine-tuned on the CombiSolv-QM dataset for 100 epochs, followed by an additional 30 epochs on the target dataset. In contrast, for SSI datasets with fewer than 3000 molecular pairs, the fine-tuning stage is shortened to 20 epochs. Notably, both pretraining and fine-tuning phases share the same optimizer and learning rate scheduling configurations, as described in the following section.

\noindent \textbf{Training Strategy.}
We employ the AdamW optimizer with a weight decay coefficient of 0.05 to mitigate overfitting and stabilize training. The learning rate is governed by a linear warm-up followed by cosine decay schedule, which helps accelerate convergence in the early stages and enables refined optimization during the later phases.
To further enhance efficiency and reduce training overhead, we adopt Low-Rank Adaptation (LoRA), implemented using the OpenDelta and PEFT libraries. The rank parameter of LoRA is set to $r=16$. For models in the DeepSeek series, LoRA is applied to the following modules: \texttt{[q\_proj, k\_proj, v\_proj, o\_proj, gate\_proj, up\_proj, down\_proj]}. For the LLaMA and Galactica models, LoRA is instead integrated into \texttt{[q\_proj, v\_proj, out\_proj, fc1, fc2]}.

\subsubsection{More Experimental Results}
\label{app:more_exp}
Due to space limitations in the main text, we present additional experimental results here, with the experimental setup consistent with the one described in the main text. 

\vspace{1em}

\begin{table*}[!htbp]
\caption{More Results of  DDI Datasets}
\label{tab:oDDI_result}

\resizebox{\textwidth}{!}{
\begin{tabular}{
  c
  >{\centering\arraybackslash}p{2.3cm}
  >{\centering\arraybackslash}p{2.3cm}
  >{\centering\arraybackslash}p{2.3cm}
  >{\centering\arraybackslash}p{2.3cm}
}
\toprule
\textbf{Experiment} & \textbf{AUC-ROC (Drugbank)} & \textbf{Accuracy (Drugbank)} & \textbf{AUC-ROC (TWOSIDES)} & \textbf{Accuracy (TWOSIDES)} \\
\midrule
Chem T5\cite{christofidellis2023unifying} & 0.921 $\pm$ 0.010 & 0.859 $\pm$ 0.013 & 0.906 $\pm$ 0.015 & 0.856 $\pm$ 0.022 \\
MolCA\cite{MolCA} & 0.934 $\pm$ 0.018 & 0.898 $\pm$ 0.010 & 0.942 $\pm$ 0.014 & 0.907 $\pm$ 0.015 \\
MolT5\cite{edwards2022translation} & 0.930 $\pm$ 0.016 & 0.904 $\pm$ 0.018 & 0.940 $\pm$ 0.013 & 0.929 $\pm$ 0.017 \\
MolTC\cite{fang2024moltc} & 0.978 $\pm$ 0.006 & 0.951 $\pm$ 0.005 & 0.980 $\pm$ 0.005 & 0.970 $\pm$ 0.007\\
\cmidrule(lr){1-5}
1.1 & 0.933 $\pm$ 0.011 & 0.891 $\pm$ 0.012 & 0.912 $\pm$ 0.017 & 0.877 $\pm$ 0.014 \\
1.2 & 0.945 $\pm$ 0.010 & 0.922 $\pm$ 0.011 & 0.957 $\pm$ 0.009 & 0.923 $\pm$ 0.009 \\
1.3 & 0.950 $\pm$ 0.009 & 0.935 $\pm$ 0.010 & 0.950 $\pm$ 0.008 & 0.926 $\pm$ 0.008 \\
1.4 & 0.955 $\pm$ 0.008 & 0.938 $\pm$ 0.009 & 0.946 $\pm$ 0.007 & 0.935 $\pm$ 0.008 \\
1.5 & 0.946 $\pm$ 0.010 & 0.931 $\pm$ 0.010 & 0.951 $\pm$ 0.008 & 0.918 $\pm$ 0.008 \\
1.6 & 0.938 $\pm$ 0.016& 0.901 $\pm$ 0.012 & 0.920 $\pm$ 0.017 & 0.893 $\pm$ 0.018 \\
1.7 & 0.963 $\pm$ 0.007 & 0.944 $\pm$ 0.007 & 0.970 $\pm$ 0.006 & 0.952 $\pm$ 0.006 \\
1.8 & 0.975 $\pm$ 0.006 & 0.950 $\pm$ 0.006 & 0.982 $\pm$ 0.005 & 0.975 $\pm$ 0.005 \\
1.9 & 0.950 $\pm$ 0.008 & 0.937 $\pm$ 0.008 & 0.971 $\pm$ 0.006 & 0.949 $\pm$ 0.006 \\
1.10 & 0.940 $\pm$ 0.010 & 0.926 $\pm$ 0.010 & 0.961 $\pm$ 0.007 & 0.938 $\pm$ 0.007 \\
1.11 & 0.920 $\pm$ 0.014 & 0.886 $\pm$ 0.010 & 0.907 $\pm$ 0.021 & 0.855 $\pm$ 0.020 \\
1.12 & 0.935 $\pm$ 0.011 & 0.920 $\pm$ 0.010 & 0.945 $\pm$ 0.012 & 0.911 $\pm$ 0.008 \\
1.13 & 0.947 $\pm$ 0.008 & 0.933 $\pm$ 0.009 & 0.953 $\pm$ 0.013 & 0.923 $\pm$ 0.009 \\
1.14 & 0.952 $\pm$ 0.007 & 0.931 $\pm$ 0.009 & 0.959 $\pm$ 0.012 & 0.932 $\pm$ 0.010 \\
1.15 & 0.935 $\pm$ 0.019 & 0.877 $\pm$ 0.020 & 0.927 $\pm$ 0.010 & 0.894 $\pm$ 0.011 \\
\cmidrule(lr){1-5}
Custom Model & \textbf{0.982 $\pm$ 0.010} & \textbf{0.956 $\pm$ 0.007} & \textbf{0.986 $\pm$ 0.007} & \textbf{0.980 $\pm$ 0.009} \\
\bottomrule
\end{tabular}
}

\end{table*}

\begin{table*}[!htbp]
\caption{More Results of  SSI Datasets}
\label{tab:oSSI_result}
\resizebox{\textwidth}{!}{
\begin{tabular}{
  c
  >{\centering\arraybackslash}p{2.3cm}
  >{\centering\arraybackslash}p{2.3cm}
  >{\centering\arraybackslash}p{2.3cm}
  >{\centering\arraybackslash}p{2.3cm}
}
\toprule
\textbf{Experiment} & \textbf{MAE \quad (MNSol)} & \textbf{RMSE (MNSol)} & \textbf{MAE (Abraham)} & \textbf{RMSE (Abraham)} \\
\midrule
Chem T5\cite{christofidellis2023unifying} & 0.537 $\pm$ 0.092 & 1.011 $\pm$ 0.083 & 0.621 $\pm$ 0.027 & 0.918 $\pm$ 0.032 \\
MolCA\cite{MolCA} & 0.511 $\pm$ 0.034 & 0.956 $\pm$ 0.049 & 0.580 $\pm$ 0.026 & 0.910 $\pm$ 0.032 \\
MolT5\cite{edwards2022translation} & 0.466 $\pm$ 0.067 & 0.867 $\pm$ 0.069 & 0.544 $\pm$ 0.028 & 0.833 $\pm$ 0.029 \\
MolTC\cite{fang2024moltc} & 0.354 $\pm$ 0.018 & 0.625 $\pm$ 0.023 & 0.211 $\pm$ 0.018 & 0.390 $\pm$ 0.021 \\
\cmidrule(lr){1-5}
1.1 & 0.510 $\pm$ 0.051 &0.971 $\pm$ 0.063 & 0.572 $\pm$ 0.024 & 0.865 $\pm$ 0.030 \\
1.2 & 0.451 $\pm$ 0.042 & 0.890 $\pm$ 0.028 & 0.524 $\pm$ 0.022 & 0.813 $\pm$ 0.027 \\
1.3 & 0.436 $\pm$ 0.018 & 0.877 $\pm$ 0.027 & 0.496 $\pm$ 0.024 & 0.804 $\pm$ 0.029 \\
1.4 & 0.410 $\pm$ 0.017 & 0.808 $\pm$ 0.024 & 0.447 $\pm$ 0.020 & 0.681 $\pm$ 0.023 \\
1.5 & 0.506 $\pm$ 0.036 & 0.944 $\pm$ 0.043 & 0.522 $\pm$ 0.030 & 0.787 $\pm$ 0.031 \\
1.6 & 0.502 $\pm$ 0.045 & 0.936 $\pm$ 0.052 & 0.591 $\pm$ 0.028 & 0.863 $\pm$ 0.025 \\
1.7 & 0.386 $\pm$ 0.034 & 0.727 $\pm$ 0.035 & 0.394 $\pm$ 0.026 & 0.512 $\pm$ 0.029 \\
1.8 & 0.343 $\pm$ 0.017 & 0.618 $\pm$ 0.024 & 0.204 $\pm$ 0.017 & 0.408 $\pm$ 0.021 \\
1.9 & 0.488 $\pm$ 0.029 & 0.834 $\pm$ 0.031 & 0.416 $\pm$ 0.027 & 0.562 $\pm$ 0.029 \\
1.10 & 0.501 $\pm$ 0.030 & 0.851 $\pm$ 0.032 & 0.456 $\pm$ 0.018 & 0.618 $\pm$ 0.022 \\
1.11 & 0.499 $\pm$ 0.016 & 0.859 $\pm$ 0.023 & 0.437 $\pm$ 0.021 & 0.620 $\pm$ 0.030 \\
1.12 & 0.550 $\pm$ 0.047 & 1.118 $\pm$ 0.064 & 0.614 $\pm$ 0.029 & 0.933 $\pm$ 0.030 \\
1.13 & 0.475 $\pm$ 0.036 & 0.878 $\pm$ 0.033 & 0.511 $\pm$ 0.019 & 0.764 $\pm$ 0.021 \\
1.14 & 0.461 $\pm$ 0.021 & 0.818 $\pm$ 0.027 & 0.489 $\pm$ 0.020 & 0.713 $\pm$ 0.021 \\
1.15 & 0.597 $\pm$ 0.031 & 0.864 $\pm$ 0.035 & 0.422 $\pm$ 0.021 & 0.683 $\pm$ 0.023 \\
\cmidrule(lr){1-5}
Custom Model & \textbf{0.340 $\pm$ 0.011} & \textbf{0.601 $\pm$ 0.010} & \textbf{0.199 $\pm$ 0.008} & \textbf{0.379 $\pm$ 0.011} \\
\bottomrule
\end{tabular}
}
\end{table*}
The results presented in Table~\ref{tab:oSSI_result} and Table~\ref{tab:oDDI_result} are largely consistent with those reported in the main text, reaffirming the trends observed across different configurations. Notably, the introduction of additional modality information consistently yields substantial improvements in model performance, highlighting the effectiveness of leveraging multimodal signals in enhancing representation learning and generalization. In contrast, architectural modifications that increase model complexity—such as altering internal modules or adding more parameters—do lead to moderate performance gains. However, these improvements are generally less pronounced compared to those achieved through the integration of new modalities. This suggests that the diversity and complementarity of multimodal data play a more critical role than mere architectural sophistication in driving performance gains.

In addition to the DDI and SSI datasets, we further evaluate our framework on the CSI dataset to demonstrate its comprehensiveness and scalability. It is worth noting that the performance of models under different configurations on the CSI dataset varies significantly. To facilitate a more intuitive comparison of the performance across different settings, we report the best-performing configuration for each backbone. The results are summarized in the Table~\ref{tab:SSI_result}. Note that the three datasets in the CSI domain are all derived by splitting the Chromophore dataset.

\begin{table*}[ht]
\caption{Performance on CSI Datasets}
\label{tab:SSI_result}
\resizebox{\textwidth}{!}{
\begin{tabular}{
  c
  >{\centering\arraybackslash}p{2.3cm}
  >{\centering\arraybackslash}p{2.3cm}
  >{\centering\arraybackslash}p{2.3cm}
  >{\centering\arraybackslash}p{2.3cm}
  >{\centering\arraybackslash}p{2.3cm}
  >{\centering\arraybackslash}p{2.3cm}
}
\toprule
\textbf{Experiment} & \textbf{MAE (Absorption)} & \textbf{RMSE (Absorption)} & \textbf{MAE (Emission)} & \textbf{RMSE (Emission)} & \textbf{MAE (Lifetime)} & \textbf{RMSE (Lifetime)}\\
\midrule
MolTC\cite{fang2024moltc} & 17.55 $\pm$ 1.83 & 29.10 $\pm$ 2.15 & 20.22 $\pm$ 1.91 & 34.17 $\pm$ 2.02 & 0.911 $\pm$ 0.052 & 1.213 $\pm$ 0.092 \\
% \cmidrule(lr){1-7}
1.4 & 18.67 $\pm$ 2.01 & 31.33 $\pm$ 2.47 & 22.37 $\pm$ 2.01 & 36.71 $\pm$ 2.93 & 1.011 $\pm$ 0.061 & 1.502 $\pm$ 0.103 \\
1.8 & 16.71 $\pm$ 1.82 & 28.67 $\pm$ 1.99 & 19.08 $\pm$ 1.89 & 38.00 $\pm$ 1.87 & 0.943 $\pm$ 0.054 & \textbf{1.119 $\pm$ 0.077} \\
1.14 & 19.01 $\pm$ 2.01 & 32.46 $\pm$ 2.92 & 21.84 $\pm$ 1.96 & 37.56 $\pm$ 3.02 & 1.009 $\pm$ 0.060 & 1.711 $\pm$ 0.095 \\
% \cmidrule(lr){1-7}
Custom Model & \textbf{15.42 $\pm$ 1.53} & \textbf{27.65 $\pm$ 1.91} & \textbf{17.11 $\pm$ 1.68} & \textbf{31.55 $\pm$ 1.60} & \textbf{0.929 $\pm$ 0.064} & 1.123 $\pm$ 0.082 \\
\bottomrule
\end{tabular}
}
\end{table*}

From the data in the Table above, we can clearly analyze the performance differences of various models under different experimental settings. Leveraging the usability and extensibility of ModuLM, we can implement and compare a wider range of LLM-based MRL models, thereby gaining insights into how model design impacts performance.
\end{document}